\documentclass[runningheads]{llncs}
\usepackage{graphicx}
\usepackage{caption}
\usepackage{subcaption}
\usepackage{svg}

\usepackage{color, soul}
\usepackage{amsfonts}
\usepackage{amsmath}
\usepackage{algorithm}
\usepackage{algorithmic}
\usepackage{booktabs}
\usepackage{multirow}
\usepackage{tabularx}
\newcolumntype{Y}{>{\centering\arraybackslash}X} 
\usepackage{amssymb}% http://ctan.org/pkg/amssymb
\usepackage{pifont}% http://ctan.org/pkg/pifont
\usepackage{appendix}
\usepackage[colorlinks=true, linkcolor=black]{hyperref}
\usepackage{footmisc}
\providecommand{\modelname}{DRFormer}

\begin{document}
\title{Dynamic Relation Transformer for Contextual Text Block Detection}
\titlerunning{Dynamic Relation Transformer for Contextual Text Block Detection}
\newcommand*\samethanks[1][\value{footnote}]{\footnotemark[#1]}
\author{
    Jiawei Wang\inst{1,3,}\thanks{Equal contribution. Correspondence to \email{wangjiawei@mail.ustc.edu.cn}, \email{shunchi.zhang@stu.xjtu.edu.cn} and \email{hk970213@mail.ustc.edu.cn}.}$^,$\samethanks[2]
    \and Shunchi Zhang\inst{2,3,}\samethanks[1]$^,$\samethanks[2]
    \and Kai Hu\inst{1,3,}\samethanks[1]$^,$\samethanks[2] \and 
    \\
    Chixiang Ma\inst{3,}\thanks{This work was done when Jiawei Wang, Shunchi Zhang, Kai Hu were interns and Chixiang Ma, Zhuoyao Zhong, Lei Sun were full-time employees in Multi-Modal Interaction Group, Microsoft Research Asia, Beijing, China.}
    \and Zhuoyao Zhong\inst{3,}\samethanks[2]
    \and Lei Sun\inst{3,}\samethanks[2]
    \and Qiang Huo\inst{3}
}
\authorrunning{Wang et al.}
\institute{
    University of Science and Technology of China \and
    Xi'an Jiaotong University \and
    Microsoft Research Asia
% \email{wangjiawei@mail.ustc.edu.cn,shunchi.zhang@stu.xjtu.edu.cn,hk970213@mail.ustc.edu.cn, } \\
% \email{chixiangma@gmail.com, zhuoyao.zhong@gmail.com,} \\ 
% \email{ray\_ustc@163.com, qianghuo@microsoft.com}
}
%

% \author{Anonymous}
% \authorrunning{Anonymous}
% \institute{Anonymous \\
% \email{Anonymous}}

\maketitle              % typeset the header of the contribution
\begin{abstract}
Contextual Text Block Detection (CTBD) is the task of identifying coherent text blocks within the complexity of natural scenes. Previous methodologies have treated CTBD as either a visual relation extraction challenge within computer vision or as a sequence modeling problem from the perspective of natural language processing. We introduce a new framework that frames CTBD as a graph generation problem. This methodology consists of two essential procedures: identifying individual text units as graph nodes and discerning the sequential reading order relationships among these units as graph edges. Leveraging the cutting-edge capabilities of DQ-DETR for node detection, our framework innovates further by integrating a novel mechanism, a Dynamic Relation Transformer (\modelname{}), dedicated to edge generation. \modelname{} incorporates a dual interactive transformer decoder that deftly manages a dynamic graph structure refinement process. Through this iterative process, the model systematically enhances the graph's fidelity, ultimately resulting in improved precision in detecting contextual text blocks. Comprehensive experimental evaluations conducted on both SCUT-CTW-Context and ReCTS-Context datasets substantiate that our method achieves state-of-the-art results, underscoring the effectiveness and potential of our graph generation framework in advancing the field of CTBD.

\end{abstract}
\section{Introduction}

% kai revised: 第一段不需要太长，介绍一下任务内容，重要性，challenges就行
Contextual Text Block Detection (CTBD) \cite{Xue22CUTE} aims to detect contextual text blocks (CTBs) within natural scenes, which are aggregates of one or more integral text units, such as characters, words, or text-lines, arranged in their natural reading order. Unlike conventional scene text detectors that primarily focus on detecting individual words or text-lines, which results in capturing fragmented textual information devoid of its full context, CTBD endeavors to capture complete and coherent text messages by detecting contextual text blocks. This is essential for subsequent tasks including natural language processing and scene image understanding. An equivalent task in document layout analysis is text region detection, which concentrates on locating text blocks within document images. However, CTBD within natural scenes presents significantly greater challenges than text region detection within document images, attributable to three principal factors: 1) The extensive diversity in text font styles and sizes encountered in natural scenes; 2) The potential lack of clear alignment among text units that comprise a single CTB; 3) The prevalence of background noises within natural scenes that can obscure text.

% Kai revised：第二段在讲text region detection的方法和局限性
The majority of existing methods in text region detection \cite{zhong2019publaynet,pfitzmann2022doclaynet,li2022dit,cheng2023m6doc,HU2024mathdetection} have been designed with a focus on document images, frequently overlooking the complex text regions encountered in natural scenes. These methods are predominantly developed for document layout analysis, relying on OCR engines, or PDF parsers to determine the bounding boxes of text units. Such approaches often neglect the essential step of detecting text units themselves, a critical component of the analysis process. Moreover, there is a prevalent practice of recognizing only physical text blocks, disregarding the concept of logical text blocks. For example, a single logical block that spans across columns is typically divided into separate physical blocks, potentially resulting in a disjointed interpretation of the document's content.

% Kai revised：第三段在讲 Contextual Text Detection 的方法和局限性
Recently, HierText dataset \cite{long2022towards} represents a significant advancement in acknowledging the hierarchical structure of text, offering a multi-level annotation schema that encompasses words, lines, and blocks in both natural scenes and document images. Despite this progress, HierText continues the trend of prior methodologies by not fully addressing the detection of logical text blocks. To address the challenge of capturing complete and meaningful text messages within the varying contexts of natural scenes,  Xue et al. \cite{Xue22CUTE} introduced two comprehensive datasets, SCUT-CTW-Context and ReCTS-Context. These datasets are purposefully crafted for the detection of logical text blocks (aka contextual text blocks). Leveraging these datasets, Xue et al. \cite{Xue22CUTE} developed Contextual Text Detector (CUTE), a method specifically designed to address contextual text block detection. CUTE adopts a natural language processing perspective to model the grouping and ordering of integral text units, extracting and transforming contextual visual features into feature embeddings to create integral text tokens, subsequently enabling the prediction of contextual text blocks. Despite these advancements, CUTE presents several limitations. Firstly, CUTE's performance is hampered in scenarios with densely packed text units, as its direct prediction of the subsequent text unit's index leads to an unwieldy and extensive index space. Secondly, the method exhibits shortcomings in precisely modeling the intricate relationships among text units that commonly arise in natural scenes. Thirdly, the utilization of an ROIAlign operator within CUTE for the extraction of local visual features proves to be insufficient for capturing a broader context required for accurate text block detection.

\begin{figure}[t]
    \centering
    \includegraphics[width=0.9\linewidth]{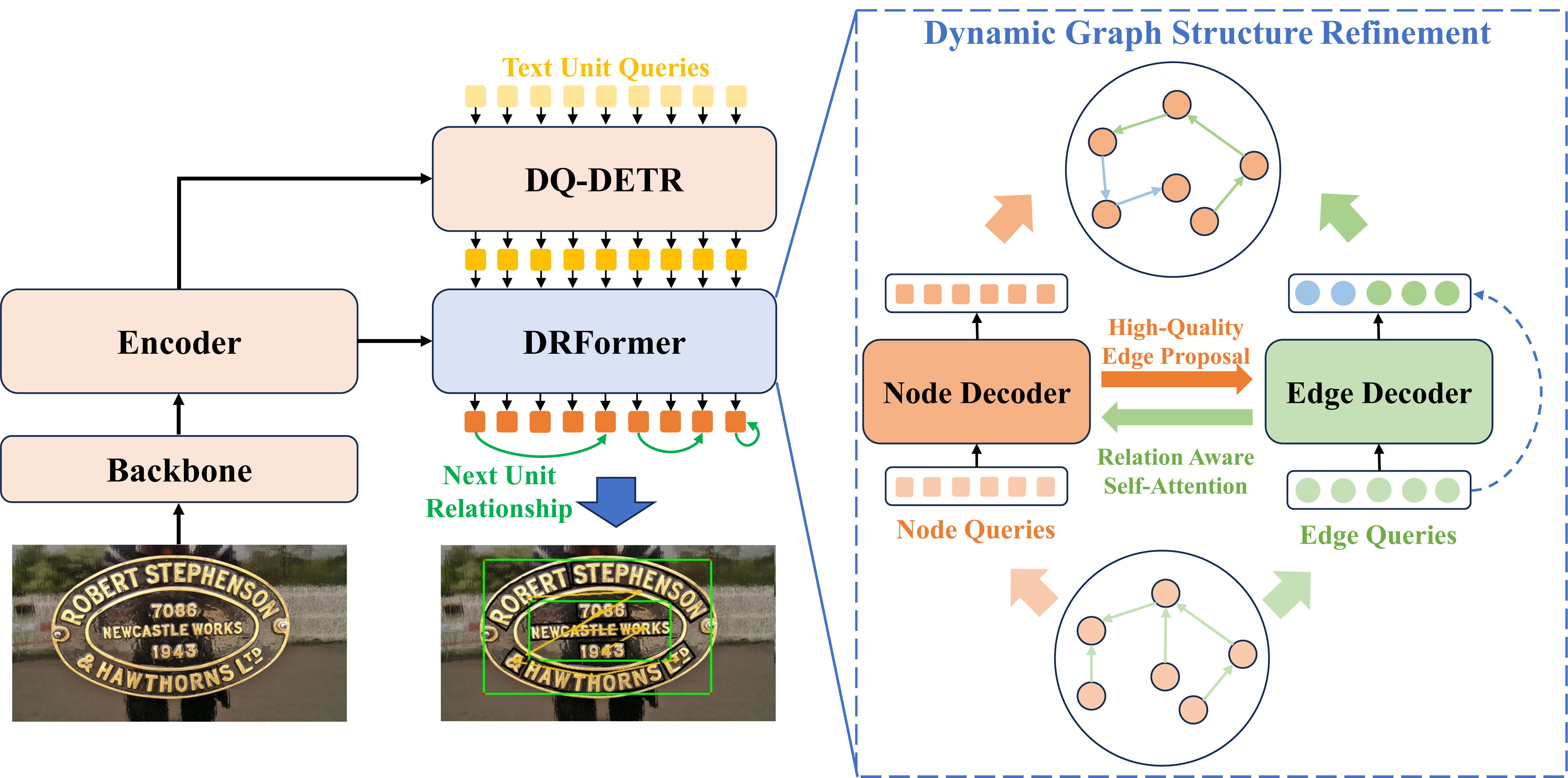}
    \caption{Overview of our proposed framework for contextual text block detection.}
    
    \label{fig:framework}
\end{figure}

% Kai revised：第四段写 我们的方法
In this paper, we introduce a novel approach to contextual text block detection by formulating it as a graph generation problem, where the graph comprises nodes and edges representing text units and the reading order relationships between them, respectively. As illustrated in Fig.~\ref{fig:framework}, our approach comprises two primary components. Firstly, a DETR-like text detector (e.g., DQ-DETR \cite{ma2023dq}), which functions as an integral text detector, is tasked with the identification of text units that constitute the graph's nodes. Secondly, a novel {\em Dynamic Relation Transformer} based relation prediction module, called \modelname{}, is dedicated to determining the reading order relationships among these detected text units, thereby facilitating the construction of the graph's edges. \modelname{} employs a dual interactive transformer decoder to
engage in a {\em dynamic graph structure refinement} process. The edge decoder enhances the node decoder's capabilities through {\em relation-aware self-attention}, which is informed by the graph's structure. Conversely, the node decoder contributes to the edge decoder's dynamic refinement process, leading to the generation of more accurate edge predictions. This iterative refinement procedure substantially improves contextual feature extraction from both node and edge queries. Moreover, we integrate a deformable attention mechanism \cite{zhu2021deformable} into our \modelname{}, which broadens its capacity to incorporate diverse contextual information effectively. The experimental results demonstrate that our proposed method achieves state-of-the-art results on SCUT-CTW-Context and ReCTS-Context datasets.

% Kai revised：第五段写 我们的contributions
The main contributions of this paper can be summarized as follows:
\begin{itemize}
    \item We are the first to propose framing the task of contextual text block detection as a graph generation problem and to introduce an iterative refinement procedure that gradually improves the quality of the generated graph.

    \item We introduce a novel relation prediction module, \modelname{}, which employs a dual interactive transformer decoder to engage in a dynamic graph structure refinement process.
    
    \item Our proposed approach has achieved state-of-the-art performance on SCUT-CTW-Context and ReCTS-Context benchmark datasets.
\end{itemize}
\section{Related Work}

% Kai revised: 
\subsection{Scene Text Detection}
Scene text detection encompasses a spectrum of methodologies, broadly categorized into bottom-up and top-down strategies. Bottom-up approaches \cite{long2018textsnake,baek2019character,zhang2020deep,ma2021relatext} typically engage object detection frameworks to initially identify discrete text components, such as characters or text segments. These text components are subsequently aggregated to construct complete text instances. Conversely, top-down methods \cite{zhong2017deeptext,zhou2017east,ma2018arbitrary,wang2019arbitrary,liu2020abcnet,zhang2022text,ma2023dq} consider entire words or text-lines as distinct object classes and employ a range of generic object detection or instance segmentation algorithms to identify them directly within a scene. For a comprehensive review of this topic, readers are referred to recent surveys in  \cite{zhu2016scene,long2021scene,naiemi2022scene}. In this paper, we adopt a state-of-the-art text detector, DQ-DETR \cite{ma2023dq}, as our text unit detection algorithm.

\subsection{Text Region Detection}
A text region is defined as a semantic unit of writing, which typically corresponds to a paragraph or a distinct block of text consisting of multiple text lines arranged according to a natural reading order. Recognizing these regions is a subtask within the broader domain of page object detection, which involves identifying and classifying various elements on a page, such as text regions, images, tables, and mathematical formulas. Text region detection can be approached through two primary methodological frameworks: bottom-up and top-down strategies.

\subsubsection{Top-Down Methods.}
These methods leverage state-of-the-art top-down object detection or instance segmentation frameworks to tackle the challenge of text region detection. Early research by Yi et al. \cite{yi2017cnn} and Oliveira et al. \cite{augusto2017fast} adapted R-CNN \cite{girshick2014rich} to identify and recognize text regions from document images. These pioneering attempts encountered limitations due to traditional strategies employed for region proposal generation. Progressing beyond these constraints, subsequent research has investigated the application of more advanced object detectors, including but not limited to Fast R-CNN \cite{girshick2015fast}, Faster R-CNN \cite{ren2015faster}, Cascade R-CNN \cite{cai2019cascade}, SOLOv2 \cite{wang2020solov2}, YOLOv5 \cite{yolov5}, and Deformable DETR \cite{zhu2021deformable} as investigated by Vo et al. \cite{vo2018ensemble}, Zhong et al. \cite{zhong2019publaynet}, Li et al. \cite{li2022dit}, Biswas et al. \cite{biswas2022docsegtr}, Pfitzmann et al. \cite{pfitzmann2022doclaynet}, and Yang et al. \cite{yang2022transformer}, respectively. Recently, it has been posited that the incorporation of textual modality information can substantially augment the efficacy of these detection frameworks. In this vein, Zhang et al. \cite{zhang2021vsr} introduced a multi-modal Faster/Mask R-CNN model for text region detection that fused visual feature maps extracted by CNN with two 2D text embedding maps containing sentence and character embeddings. Furthermore, Li et al. \cite{li2022dit}, and Huang et al. \cite{huang2022layoutlmv3} enhanced the capabilities of Faster R-CNN, Mask R-CNN, and Cascade R-CNN-based text region detectors by employing the pre-training of vision backbone networks on extensive corpora of document images via self-supervised learning algorithms. Although top-down methods have achieved significant success, they fall short in text region detection in natural scenes where text units within a same region can be widely dispersed and different regions may significantly overlap, presenting substantial challenges for accurate detection.

\subsubsection{Bottom-Up Methods.}
These approaches formulate text region detection as a relation prediction problem, aiming to group text units (e.g., words, text-lines, connected components) into text regions by predicting relationships between them. Various approaches (e.g., \cite{li2018page,li2020page,luo2022doc,wang2022post}) represent each document page as a graph, where nodes correspond to basic text units, and edges denote inter-unit relationships. For instance, Li et al. \cite{li2018page} employed image processing techniques to initially generate line regions, subsequently employing two CRF models to predict whether pairs of line regions belong to a same text region, based on visual features extracted by CNNs. Advancing their research, Li et al. \cite{li2020page} substituted line regions with connected components as graph nodes and utilized a graph attention network (GAT) to refine the visual features of both nodes and edges. Wang et al. \cite{wang2022post} focused on paragraph identification, developing a GCN-based approach to cluster text-lines into coherent paragraphs. Liu et al. \cite{liu2022unified} introduced a unified framework designed to concurrently perform text detection and paragraph (text-block) identification. Furthermore, Zhong et al. \cite{zhong2023hybrid} advanced a multi-modal transformer-based model for the prediction of relations, thereby facilitating text region detection.

Despite achieving state-of-the-art results on several benchmark datasets, these methods predominantly target text region detection within document images, thereby neglecting the intricately structured text regions commonly found in natural scenes. Recent endeavors to bridge this gap have led to the introduction of datasets specifically designed for text region detection in natural scenes, such as HierText \cite{long2022towards}, SCUT-CTW-Context \cite{Xue22CUTE}, and ReCTS-Context \cite{Xue22CUTE}. Long et al. \cite{long2022towards} adopted an end-to-end instance segmentation model to detect arbitrarily shaped text and incorporated a multi-head self-attention layer to aggregate text regions. Xue et al. \cite{Xue22CUTE} approached text region detection from an NLP standpoint, conceptualizing text units as tokens that are subsequently aggregated into coherent token sequences corresponding to a same text region. Nevertheless, these approaches exhibit suboptimal performance in the presence of numerous text units. In this paper, we are the first to propose framing the task of text region detection within natural scenes as a graph generation problem and introducing an iterative refinement procedure that progressively improves the quality of the generated graphs.
% kai revised
\section{Methodology}

Our method conceptualizes the task of contextual text block detection as a graph generation problem, comprising two key components: (1) An integral text detector based on DQ-DETR \cite{ma2023dq}, tasked with identifying individual text units and representing them as nodes within a graph; (2) A {\em dynamic relation transformer}, designed to discern the reading order relationships among these text units, thereby establishing the edges of the graph.  These two components are jointly trained in an end-to-end manner. The overall architecture of our proposed approach is illustrated in Fig.~\ref{fig:framework}. Subsequent subsections provide a detailed exposition of these components.

\subsection{Integral Text Detector}
\label{sec:text-unit-detection}
Given an input image, we leverage a CNN backbone, such as ResNet \cite{he2016deep}, to generate a set of three multi-scale feature maps $\{C_3, C_4, C_5\}$. These feature maps are subsequently processed by a deformable transformer encoder \cite{zhu2021deformable}, which enriches the pixel-level embeddings across each feature map. Leveraging the enhanced feature representations, we employ the DQ-DETR decoder to identify all text units of arbitrary shapes within the image, denoted by $\{t_1, t_2, t_3, ..., t_n\}$. For an in-depth understanding of the DQ-DETR framework, readers may consult the original publication \cite{ma2023dq}. The detected text units are then conceptualized as the nodes of a graph. These nodes are input into our dynamic relation transformer, which is tasked with predicting the reading order relationships among them, thus forming the graph's edges.

\subsection{Dynamic Relation Transformer}

\begin{figure}[t]
    \centering
    \includegraphics[width=0.9\linewidth]{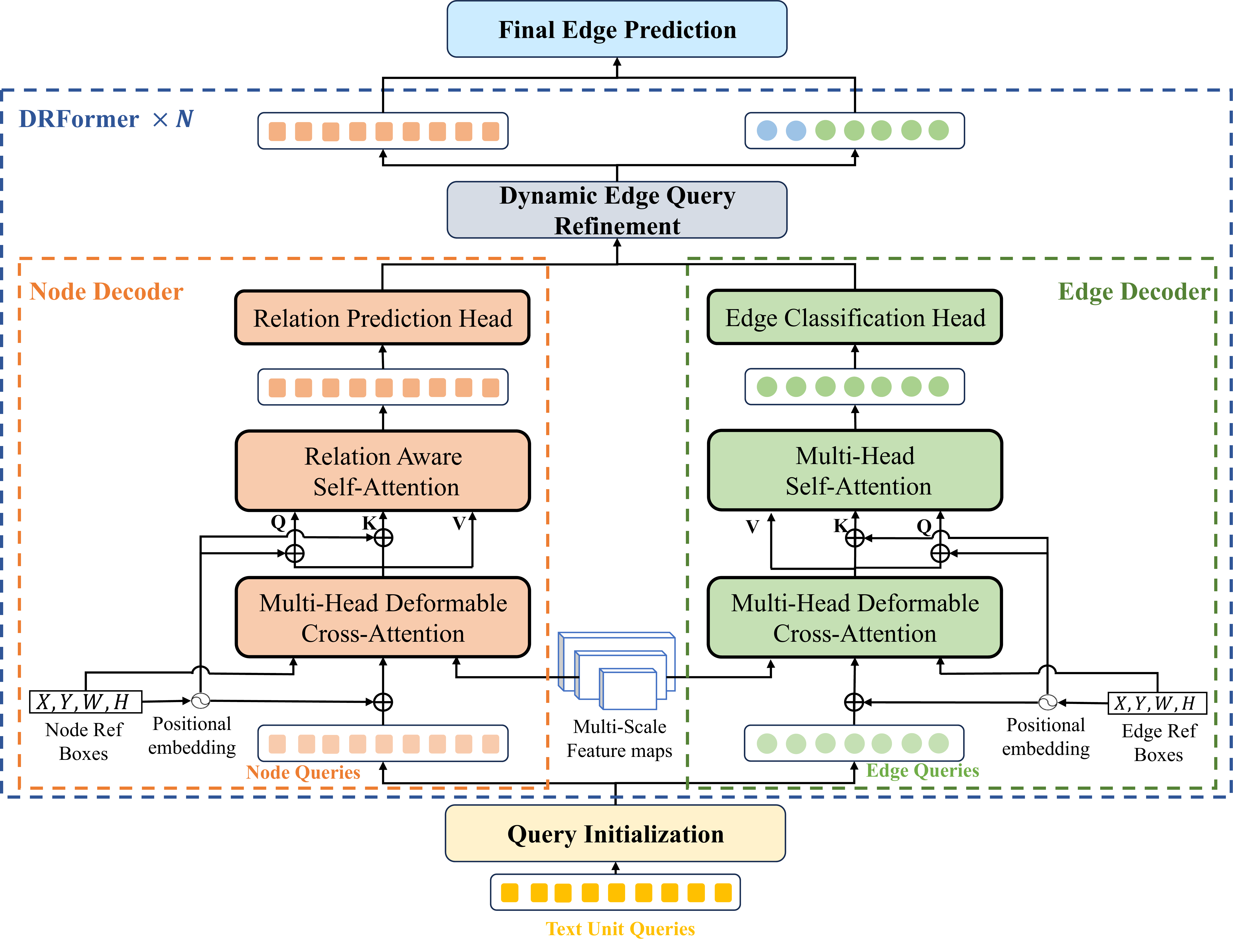}
    \caption{The architecture of our proposed \modelname{}, consisting of a dual interactive decoder. For clarity, we only show two attention layers of Transformer decoder and omit the FFN blocks.}
    \label{fig:dreamer}
\end{figure}

As depicted in Fig.~\ref{fig:dreamer}, our \modelname{} introduces a {\em dynamic graph structure refinement} process, which iteratively enhances the quality of the generated graph. The process begins by generating initial node and edge queries through a {\em query initialization} module. These initial queries are then input to a {\em dual interactive decoder}, tasked with augmenting their embeddings. Following this, a relation prediction head and an edge classification head collaboratively work to refine the connections within the graph. Based on their output, incorrect edges are pruned and potential edge queries are formulated and added to the next decoder layer to assist in identifying further connections for the graph. Ultimately, the refined node and edge query embeddings produced by the terminal decoder layer are passed to the edge classification head, which plays a crucial role in filtering out false positives and in determining the correct reading order relationships among all the text units in the input image.

\subsubsection{Query Initialization.}
For the node queries, we utilize the query embeddings and the bounding boxes of the detected text units produced by the DQ-DETR as the node embeddings $\{q^N_1, q^N_2, ..., q^N_n\}$ and corresponding node reference boxes $\{B^N_1, B^N_2, ..., B^N_n\}$. Regarding the edge queries, we adopt the relation prediction head as proposed by Zhong et al. \cite{zhong2023hybrid} to predict the subsequent text unit for each identified text unit. To bolster the recall rate of edge proposals, we select the top-K potential successors $\{t_{i_1}, t_{i_2}, ..., t_{i_k}\}$ for each text unit $t_i$, thereby establishing the initial edge proposals. The initialization of edge queries $\{q^E_{11}, q^E_{12}, ..., q^E_{nk}\}$ and their respective edge reference boxes $\{B^E_{11}, B^E_{12}, ..., B^E_{nk}\}$ is formalized as follows:
\begin{gather}
\label{edge_init}
q^E_{ij} = FC_{init}([q^N_i, q^N_{i_j}]), \\
B^E_{ij} = Union\_Box([B^N_i, B^N_{i_j}]),
\end{gather}
where $FC_{init}$ denotes a dedicated fully-connected layer designed to merge the embeddings of $q^N_i$ and $q^N_{i_j}$, and $Union\_Box$ is a function that computes the union bounding rectangle for the pair of bounding boxes $B^N_i$ and $B^N_{i_j}$.

\subsubsection{Dual Interactive Decoder.}

\label{sec:dual_inter_decoder}
As illustrated in Fig.~\ref{fig:dreamer}, our \modelname{} includes a dual interactive decoder, consisting of a node decoder and an edge decoder, each dedicated to refining the embeddings of their respective graph components. Inspired by the Mask2Former \cite{cheng2022masked}, our dual interactive decoder deviates from the vanilla deformable decoder architecture \cite{zhu2021deformable} by resequencing the cross-attention and self-attention operations within the decoder layers. The decoding process begins with the application of a deformable cross-attention module that dynamically samples relevant image features from the encoder's multi-scale feature maps. Following this, a self-attention mechanism is engaged to model the interactions among nodes and edges. To further concentrate the node decoder's attention on the complex relationships inherent to the graph structure, we incorporate a {\em relation-aware self-attention} module into the node decoding workflow. This novel module adheres to a guiding principle: it enables attention exchange exclusively among node pairs that are part of the same undirected connected sub-graph, while simultaneously restricting attention flow across nodes that belong to disparate connected sub-graphs. By employing this dual interactive attention approach, our system refines both node and edge embeddings with increased precision, thereby enhancing the overall accuracy of the graph generation.

\subsubsection{Dynamic Edge Query Refinement.}
The edge query embeddings, updated at each iteration of the edge decoder, are input to a relation classification head designed to filter out incorrect edge proposals generated by the preceding layer. This classification head operates as a binary classifier, where proposals receiving low edge classification scores are pruned from the set of edge proposals. Post this filtration step, certain nodes may become isolated, devoid of any connecting edges. To address this, the enhanced node query embeddings from the node decoder are fed into a relation prediction head, which is tasked with generating the top-K most likely edges to potentially connect these isolated nodes. These candidate edges are then incorporated into the set of edge proposals as new edge queries for the subsequent decoder layer. Through this recursive refinement process, each successive layer of the decoder contributes to progressively improving the accuracy of the relationship prediction within the graph.

\subsubsection{Final Edge Prediction.}
The node query embeddings and edge query embeddings, emitted by the terminal decoder layer, are channeled into both a relation prediction head and a relation classification head, which are responsible for generating the definitive edges of the resultant graph. In terms of architecture, both the relation prediction head and the relation classification head mirror their counterparts from the preceding layers.

\subsection{Optimization}
\label{sec:opt}

\subsubsection{Loss for Integral Text Detector.}
The loss function for our detector, denoted as $L_{detector}$, aligns precisely with the loss function $L_{DQ-DETR}$ presented in \cite{ma2023dq}. This composite loss function consolidates a set of bounding box regression losses and classification losses stemming from the diverse prediction heads and denoising heads within the architecture. Specifically, the bounding box regression loss is formulated as a weighted sum of the $L_1$ loss and the GIoU loss \cite{detr2020}, while the classification loss employs the focal loss \cite{lin2017focal} for its calculation. For a comprehensive understanding of these loss components, we direct the readers to the detailed descriptions provided in \cite{ma2023dq}.

\subsubsection{Loss for Relation Prediction Head.}
During the initial phase of edge query generation and at each stage of the decoder, we employ a relation prediction head to formulate edge proposals for the subsequent layer of the decoder. A consistent softmax cross-entropy loss is utilized across all relation prediction heads, which can be expressed as follows:
\begin{equation}
    L^{(l)}_{relation} = \frac{1}{N} \sum_{i} L_{\mathrm{CE}}\left(\boldsymbol{s}_{i}, \boldsymbol{s}_{i}^*\right)
\end{equation}
where $\boldsymbol{s}_{i}$ denotes the predicted relation score vector that encapsulates the probability distribution for the text unit $t_i$ linking to other text units, while $\boldsymbol{s}_{i}^*$ represents the corresponding ground truth label.

\subsubsection{Loss for Edge Classification.}
For the task of edge classification, we employ a binary cross-entropy loss, which is articulated as follows:
\begin{equation}
    L^{(l)}_{edge} = \frac{1}{N} \sum_{i} L_{\mathrm{BCE}}\left(\boldsymbol{c}_{i}, \boldsymbol{c}_{i}^*\right)
\end{equation}
where $\boldsymbol{c}_{i}$ signifies the logits corresponding to the $i$-th edge proposal as produced by the edge classification head, while $\boldsymbol{c}_{i}^*$ denotes the associated ground-truth label.

\subsubsection{Overall Loss.} 
All the components in our approach are jointly trained in an end-to-end manner. The overall loss is the sum of $L_{detector}$, $L^{(l)}_{relation}$ and $L^{(l)}_{edge}$:
\begin{equation}
    L_{overall} = L_{detector} + \sum_{l=0}^{L} L^{(l)}_{relation} + \sum_{l=1}^{L} L^{(l)}_{edge}
\end{equation}
where $L$ signifies the total number of decoder layers in the architecture, with $l=0$ representing the initial phase of edge query generation, and $l=1$ to $L$ corresponding to the subsequent decoder layers.
\section{Experiments}

\subsection{Datasets}

We conduct our experiments on SCUT-CTW-Context and ReCTS-Context datasets, both introduced in \cite{Xue22CUTE}, to validate the effectiveness of our proposed framework.

\textbf{SCUT-CTW-Context:} The dataset, as annotated by Xue et al. \cite{Xue22CUTE}, augments SCUT-CTW-1500 dataset \cite{yuliang2017detecting} with contextual text blocks. It consists of a corpus of 940 training images and 498 test images. The majority of integral text units in this dataset are words, providing rich contextual information captured in various scenes.

\textbf{ReCTS-Context:} Similarly annotated by Xue et al. \cite{Xue22CUTE}, this dataset enhances ICDAR2019-ReCTS \cite{liu2019icdar} with reading order relationship annotations. It is partitioned into a training set of 15,000 images and a test set comprising 5,000 images. Characterized by Chinese script, the dataset presents characters as the fundamental textual elements, offering a unique challenge in the realm of reading order relationship prediction.

\subsection{Evaluation Metrics}

We adopt evaluation metrics proposed in \cite{Xue22CUTE} for the assessment of contextual text detection, which includes {\em Local Accuracy}, {\em Local Continuity}, and {\em Global Accuracy}. These metrics provide a comprehensive evaluation of the effectiveness of the proposed framework.

\textbf{Local Accuracy (LA):} This metric is employed to assess the accuracy of order prediction for neighboring integral text units, focusing on the local characteristics of integral text unit ordering.

\textbf{Local Continuity (LC):} LC evaluates the continuity of integral text units by computing a modified $n$-gram precision score, inspired by BLEU \cite{papineni2002bleu}. Similar to LA, LC also focuses on the local characteristics of integral text unit ordering.

\textbf{Global Accuracy (GA):} GA is introduced to evaluate the detection accuracy of contextual text blocks, representing a stringent constraint that provides a comprehensive measure of the overall performance of the proposed framework.

Moreover, a detected integral text unit is considered matched with a ground-truth text unit if the intersection-over-union (IoU) of their bounding boxes exceeds a certain threshold. We employ three commonly used IoU threshold standards in generic object detection tasks, including $IoU = 0.5$, $IoU=0.75$, and $IoU = 0.5 : 0.05 : 0.95$ for comprehensive evaluation. 

\subsection{Implementation Details}
Our methodology is implemented using PyTorch, and the experiments are conducted on a workstation equipped with 16 NVIDIA Tesla V100 GPUs (32 GB memory). To achieve the highest possible performance, we employ the latest state-of-the-art text detector, namely DQ-DETR \cite{ma2023dq}, as the integral text detector in our framework. We use the same configuration as in DQ-DETR. The backbone in our framework is initialized with ResNet-50, which is pre-trained on ImageNet-1K dataset. The node decoder and edge decoder in \modelname{} are both configured with 3 layers. Both are designed with the number of heads, the dimension of the hidden state, and the dimension of the feedforward network set as 8, 256, and 1024, respectively. The number of edge proposals for each node, denoted as $K$, is set to 3. 

In accordance with the training strategy proposed in CUTE \cite{Xue22CUTE}, our approach begins by training the integral text detector on the SCUT-CTW-Context and ReCTS-Context datasets separately. We adopt the same training configuration as DQ-DETR, with the only difference being that we do not pre-train our model on a mixture of SynthText150K \cite{liu2020abcnet}, MLT2017 \cite{nayef2017icdar2017} and Total-Text \cite{ch2017total}. Subsequently, we proceed to train the overall model while freezing the parameters in the backbone and DQ-DTER. To optimize the models, we employ Adam algorithm \cite{kingma2014adam} with the following settings: a learning rate of 5e-4, betas set to (0.9, 0.999), epsilon set to 1e-8, and weight decay set to 1e-2. All models are trained for 5,000 iterations with a warmup strategy applied during the first 200 iterations. In each training iteration, we select 8 samples as a mini-batch for each GPU. We employ a similar multi-scale training strategy in DQ-DETR, randomly resizing shorter sides of an image to sizes ranging from 640 to 2,560 pixels without keeping aspect ratios. During inference, the longer side of each testing image is set to 1600.

\subsection{Comparisons with Prior Arts}

In this section, we conduct a comprehensive comparative analysis between \modelname{} and several methods proposed in \cite{Xue22CUTE} across the SCUT-CTW-Context and ReCTS-Context datasets. To underscore the effectiveness of our proposed framework, we construct a solid baseline method, denoted as Baseline. The baseline configuration mirrors the structural pipeline of our framework, with the exception that it substitutes the \modelname{} with a simpler 3-layer node decoder. Structurally, this baseline is akin to the text region detection architecture discussed in \cite{zhong2023hybrid}, with distinctive modifications. It diverges by replacing the multi-modal transformer encoder with a deformable decoder and by only leveraging visual modality information.

\subsubsection{SCUT-CTW-Context.}
We benchmark \modelname{} against existing baseline models on the SCUT-CTW-Context dataset, where integral text units are delineated at the granularity of individual words. Table~\ref{table:main-ctw} demonstrates that our robust Baseline model, which incorporates our proposed graph generation framework, exhibits marked superiority by outperforming the LINK-R101 \cite{xue2022detection} and CUTE-R101 \cite{Xue22CUTE} models. This showcases the effectiveness of our framework in the relevant domain. Building upon this robust foundation, our proposed \modelname{} employs a {\em dynamic graph structure refinement} process to further enhance the accuracy of reading order relationships. This strategic refinement leads to \modelname{} significantly outperforming the strong Baseline, achieving noteworthy improvements across all metrics on the SCUT-CTW-Context dataset. Specifically, at an IoU threshold of 0.5, \modelname{} achieves a 2\% increase in Local Accuracy (LA), a 3.3\% rise in Local Continuity (LC), and a 2\% enhancement in Global Accuracy (GA). Notably, this superior performance is sustained even when faced with more stringent IoU thresholds. Within the challenging IoU evaluation spectrum of 0.5 to 0.95, incremented by 0.05, \modelname{} continues to outpace the Baseline, showing increments of 1.2\% in LA, 2.2\% in LC, and 1.4\% in GA, thereby confirming the model's robustness and consistent outperformance against more rigorous benchmarks.

\begin{table}[t]

\caption{Quantitative performance comparison of \modelname{} with state-of-the-art methods on SCUT-CTW-Context dataset. LA: Local Accuracy; LC: Local Continuity; GA: Global Accuracy.}

\begin{center}
% \resizebox{\linewidth}{!}{%
\begin{tabular}{ll ccc ccc ccc}
\toprule
& \multirow{2}{*}{Models}
& \multicolumn{3}{c}{IoU=0.5}
& \multicolumn{3}{c}{IoU=0.75}
& \multicolumn{3}{c}{IoU=0.5:0.05:0.95} \\
\cmidrule(lr){3-5} \cmidrule(lr){6-8} \cmidrule(lr){9-11}
&
& ~~LA~~ & ~~LC~~ & ~~GA~~
& ~~LA~~ & ~~LC~~ & ~~GA~~
& ~~LA~~ & ~~LC~~ & ~~GA~~ \\
\midrule
% & CLUSTERING      & 18.4 & 7.9  & 6.8  & 14.1 & 5.9  & 4.7  & 13.5 & 5.7  & 4.9  \\
& LINK-R50 \cite{xue2022detection} & 25.5 & 3.3  & 18.9 & 20.3 & 3.2  & 14.7 & 19.3 & 2.9  & 14.3 \\
& CUTE-R50 \cite{Xue22CUTE}        & 54.0 & 39.2 & 30.7 & 41.6 & 31.2 & 23.7 & 39.4 & 29.0 & 22.1 \\
& LINK-R101 \cite{xue2022detection} & 25.7 & 3.4  & 19.2 & 20.0 & 2.9  & 14.7 & 19.6 & 2.7  & 14.4 \\
& CUTE-R101 \cite{Xue22CUTE}        & 55.7 & 39.4 & 32.6 & 40.6 & 29.0 & 22.8 & 40.0 & 28.3 & 22.7 \\
\midrule
& Baseline-R50 & 67.6 & 55.7	& 45.8 & 56.5 & 43.6 & 37.3 & 47.4 & 37.1 & 31.9 \\
& \modelname{}-R50  & \textbf{69.6} & \textbf{59.0} & \textbf{47.8} & \textbf{58.1} & \textbf{46.0} & \textbf{39.3} & \textbf{48.9} & \textbf{39.3} & \textbf{33.3} \\
\bottomrule
\end{tabular}
% }
\end{center}
\label{table:main-ctw}
\end{table}
\begin{table}[t]

\caption{Quantitative performance comparison of \modelname{} with state-of-the-art methods on ReCTS-Context dataset. LA: Local Accuracy; LC: Local Continuity; GA: Global Accuracy.}

\begin{center}
% \resizebox{\linewidth}{!}{%
\begin{tabular}{ll ccc ccc ccc}
\toprule
& \multirow{2}{*}{Models}
& \multicolumn{3}{c}{IoU=0.5}
& \multicolumn{3}{c}{IoU=0.75}
& \multicolumn{3}{c}{IoU=0.5:0.05:0.95} \\
\cmidrule(lr){3-5} \cmidrule(lr){6-8} \cmidrule(lr){9-11}
&
& ~~LA~~ & ~~LC~~ & ~~GA~~
& ~~LA~~ & ~~LC~~ & ~~GA~~
& ~~LA~~ & ~~LC~~ & ~~GA~~ \\
\midrule
% & Clustering                   & 32.2 & 19.1 & 10.6 & 26.1 & 17.0 & 9.7  & 25.6 & 16.1 & 9.0  \\
& LINK-R50 \cite{xue2022detection} & 68.2 & 57.5 & 48.4 & 53.8 & 50.2 & 38.4 & 53.0 & 47.7 & 37.3 \\
& CUTE-R50 \cite{Xue22CUTE}        & 70.4 & 64.7 & 51.6 & 54.4 & 56.6 & 39.5 & 53.9 & 53.6 & 38.9 \\
& LINK-R101 \cite{xue2022detection} & 70.8 & 59.1 & 49.9 & 54.5 & 51.0 & 39.0 & 53.4 & 48.3 & 37.9 \\
& CUTE-R101 \cite{Xue22CUTE}        & 72.4 & 67.3 & 53.8 & 55.1 & \textbf{57.0} & 40.2 & 54.6 & \textbf{53.9} & 39.4 \\
\midrule
& Baseline-R50              & 82.2 & 71.4 & 69.6 & 63.2 & 50.0 & 52.8 & 56.4 & 46.0 & 47.6 \\
& \modelname{}-R50            & \textbf{83.3} & \textbf{74.6} & \textbf{71.8} & \textbf{67.6} & 55.9 & \textbf{56.9} & \textbf{59.4} & 50.0 & \textbf{50.6} \\
\bottomrule
\end{tabular}
% }
\end{center}
\label{table:main-rects}
\end{table}

\subsubsection{ReCTS-Context.}
Moreover, the effectiveness of our \modelname{} is convincingly validated on the ReCTS-Context dataset, which is specifically designed to extract complex Chinese text displayed on signboards. The layout and arrangement of Chinese characters on signboards present unique challenges, differing significantly from other scenes. In the face of such intricacies, \modelname{} demonstrates its superior performance capabilities, particularly within the stringent IoU range of 0.5 to 0.95. As summarized in Table~\ref{table:main-rects}, within this rigorous evaluative context, \modelname{} delivers a noteworthy 3\% improvement in Local Accuracy, a significant 4\% enhancement in Local Continuity, and a remarkable 2.9\% augmentation in Global Accuracy, surpassing the strong Baseline and solidifying its adeptness at handling complex text arrangements on signboards.

\begin{table}[t]

\caption{Quantitative performance comparison of \modelname{} with state-of-the-art methods on integral text grouping and ordering task on SCUT-CTW-Context and ReCTS-context datasets.}

\begin{center}
\begin{tabular}{ll ccc ccc}
\toprule
& \multirow{2}{*}{Models}
& \multicolumn{3}{c}{SCUT-CTW-Context}
& \multicolumn{3}{c}{ReCTS-Context} \\
\cmidrule(lr){3-5} \cmidrule(lr){6-8}
&
& ~~LA~~ & ~~LC~~ & ~~GA~~
& ~~LA~~ & ~~LC~~ & ~~GA~~ \\
\midrule
& LINK-R50 & 30.2 & 4.5 & 22.8 & 83.8 & 68.4 & 61.1 \\
& CUTE-R50 & 71.5 & 58.5 & 49.7 & 92.1 & 82.8 & 76.0 \\
& LINK-R101 & 45.5 & 6.3 & 31.7 & 86.7 & 75.0 & 69.6 \\
& CUTE-R101 & 71.5 & 58.7 & 52.6 & \textbf{93.1} & 83.7 & 77.8 \\
\midrule
& Baseline-R50 & 80.3 & 71.0 & 58.7 & 90.9 & 81.8 & 82.8 \\
& \modelname{}-R50 & $\mathbf{83.9}$ & $\mathbf{76.0}$ & $\mathbf{60.5}$ & $92.8$ & $\mathbf{85.9}$ & $\mathbf{85.5}$ \\
\midrule
\end{tabular}
\end{center}
\label{table:gt}
\end{table}

\subsubsection{Upper Bound Evaluation with GT Text Units.}
To validate the effectiveness of \modelname{} solely in grouping and ordering integral text units, we introduce an additional evaluation criterion that assumes the accurate detection of all integral text units by leveraging the bounding boxes of ground-truth integral text units. As shown in Table~\ref{table:gt}, the proposed \modelname{} demonstrates enhanced capability in the grouping and ordering of integral text units, outperforming all baseline models in this respect.

\subsection{Ablation Studies}

To thoroughly evaluate the effectiveness of various design aspects within \modelname{} for the grouping and ordering of integral text units, we conducted a series of ablation experiments on the SCUT-CTW-Context dataset. These experiments assume the accurate detection of all integral text units by leveraging the bounding boxes of ground-truth integral text units.

\subsubsection{Dynamic Graph Structure Refinement.} 
Building upon a robust baseline model, our novel \modelname{} integrates a dynamic graph structure refinement mechanism, which incrementally optimizes the quality of the constructed graph. This is achieved through the implementation of a dual interactive transformer decoder that facilitates a synergistic interaction between nodes and edges during the refinement process. As evidenced by the data presented in rows 1 and 2 of Table~\ref{table:ablation}, the incorporation of the {\em Dynamic Graph Structure Refinement} leads to a measurable performance boost, with a 2\% gain in Local Accuracy (LA) and a 1.6\% increase in Local Continuity (LC). These statistics clearly demonstrate the critical contribution of our dynamic graph structure refinement approach in effectively predicting reading order relations within the generated graph.

\begin{table}[t]
\caption{Ablation studies of various components within \modelname{} on SCUT-CTW-Context dataset. (DGSR: Dynamic Graph Structure Refinement; CAF: Cross-Attention First; RASA: Relation-Aware Self-Attention)}
\centering
\begin{tabularx}{0.8\textwidth}{cYYYYYYY}  
\toprule
\# & Method & DGSR & CAF & RASA & LA & LC & GA \\
\midrule
1 & Baseline & & & & 80.3 & 71.0 & 58.7 \\
\midrule
2 & & \checkmark & & & 82.3 & 72.6 & 58.8 \\
3 & & \checkmark & \checkmark & & 83.4 & 75.3 & 60.2 \\
\midrule
4 & \modelname{} & \checkmark & \checkmark & \checkmark & \textbf{83.9} & \textbf{76.0} & \textbf{60.5} \\
\bottomrule
\end{tabularx}  
\label{table:ablation}
\end{table}

\subsubsection{Order of Self-Attention and Cross-Attention.} 
As outlined in Section~\ref{sec:dual_inter_decoder}, we propose reversing the conventional order of self-attention and cross-attention mechanisms within the transformer decoder architecture. In the context of our task, which involves predicting reading order relationships among text units, it is beneficial for the relation prediction model to first engage the cross-attention module to distill features from text units. Subsequently, the incorporation of the self-attention module empowers the model to skillfully decipher the interactions among these units, thereby augmenting its ability for accurate relationship prediction. The empirical results, as indicated in rows 2 and 3 of Table~\ref{table:ablation}, reveal that the adoption of the {\em Cross-Attention First} strategy leads to a notable performance improvement, evidenced by a 1.1\% rise in LA and a substantial 2.7\% surge in LC.

\subsubsection{Relation-Aware Self-Attention.}
Within the structure of our proposed \modelname{}, the edge decoder collaborates closely with the node decoder, imparting an understanding of the graph's structure. This collaboration empowers the node decoder to utilize {\em Relation-Aware Self-Attention}, enriching its comprehension of the intra-graph relationships. The outcomes of our ablation study, as shown in rows 3 and 4 of Table~\ref{table:ablation}, indicate a performance enhancement when {\em Relation-Aware Self-Attention} is employed instead of the conventional self-attention mechanism. Specifically, we observe an increment of 0.5\% in LA and 0.7\% in LC, underscoring the benefits of our {\em Relation-Aware Self-Attention}.

\subsection{Qualitative Results}

\begin{figure}[ht]
    \centering
    \begin{subfigure}[c]{\textwidth}
        \includegraphics[width=\textwidth]{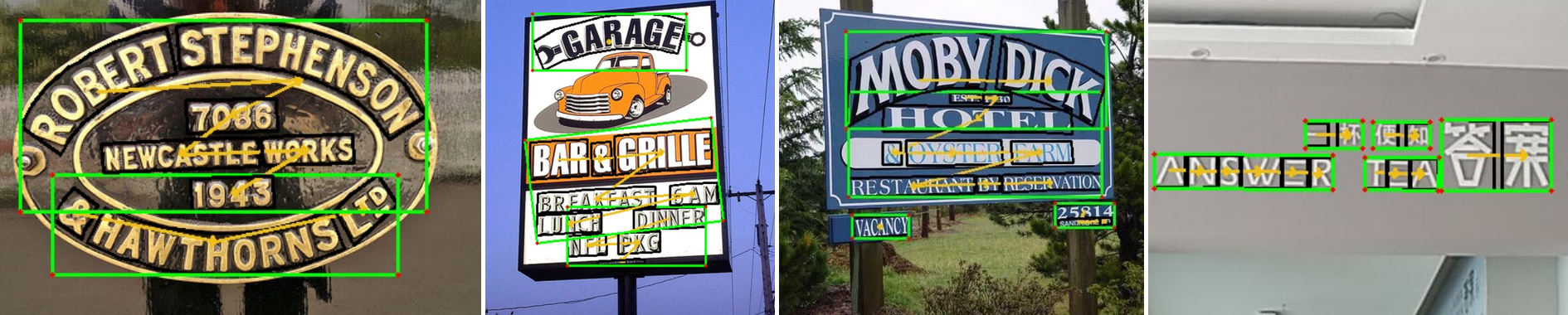}
    \end{subfigure}
    \begin{subfigure}[c]{\textwidth}
        \includegraphics[width=\textwidth]{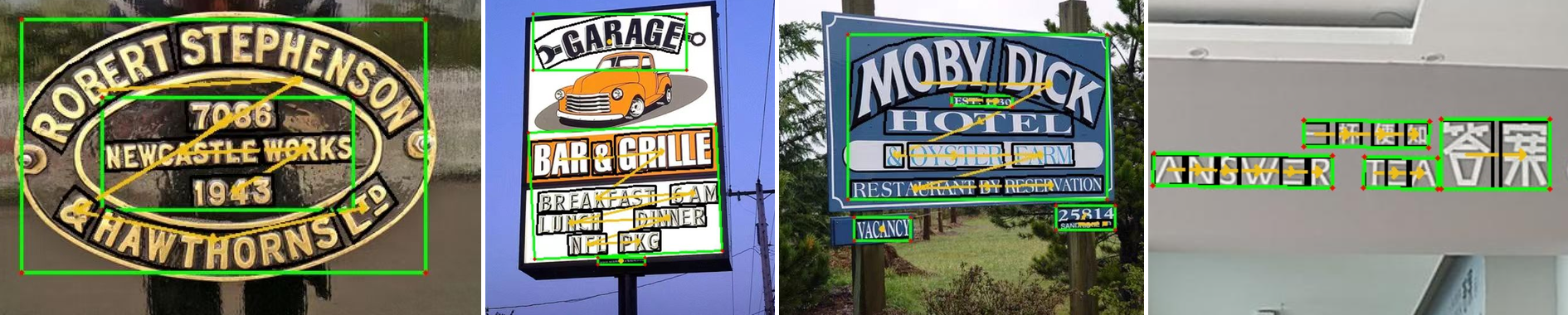}
    \end{subfigure}
    \caption{Qualitative comparison between our proposed baseline (top) and \modelname{} (bottom). Black bounding boxes indicate word-level or character-level integral texts, \textit{i.e.}, graph's nodes, and brown arrows represent the reading order relationships between these integral texts, \textit{i.e.}, graph's edges, which finally lead to contextual text blocks outlined in green bounding boxes. Best viewed in color.}
    % \caption{Qualitative comparison between our proposed baseline (top) and \modelname{} (bottom). {\color{black}\textbf{Black}} bounding boxes indicate word-level or character-level integral texts, \textit{i.e.}, graph's nodes, and {\color{brown}\textbf{brown}} arrows represent the sequential relationships between these integral texts, \textit{i.e.}, graph's edges, which finally lead to contextual text blocks outlined in {\color{green}\textbf{green}} bounding boxes. Best viewed in color.}
    \label{fig:qualitative}
\end{figure}

To emphasize the effectiveness of our proposed methods, we offer a qualitative comparison between our baseline and \modelname{}. As depicted in Fig.~\ref{fig:qualitative}, \modelname{} demonstrates superior accuracy in establishing edge relationships through dynamic graph structure refinement, significantly improving the overall quality of contextual text block detection compared to our baseline method.

\section{Conclusion and Future Work}

In this paper, we introduce a novel framework for addressing the task of contextual text block detection by formulating it as a graph generation problem. To demonstrate the effectiveness of our framework, we propose {\em Dynamic Relation Transformer} (\modelname{}), a novel relation prediction module that integrates the graph structure into the relation prediction process. Additionally, \modelname{} introduces a {\em Dynamic Graph Structure Refinement} procedure through the incorporation of a dual interactive transformer decoder. This iterative process progressively augments the graph's fidelity, culminating in an enhanced precision of contextual text block demarcation. The promising results obtained in this study suggest that our proposed framework opens up promising avenues for future research in the field of scene text region detection.

In the future, we will explore the role of contextual text blocks in text spotting and introduce text embedding to assist in better relationship prediction. Additionally, we aim to explore the integration of the {\em Dynamic Graph Structure Refinement} concept into other tasks related to relationship prediction or graph generation.
\bibliographystyle{splncs04}
\bibliography{bibfile}

\end{document}